\pdfoutput=1
\documentclass[10pt]{article}

\usepackage[preprint, nonatbib]{neurips_2026}
\usepackage[T1]{fontenc}
\usepackage[utf8]{inputenc}
\DeclareUnicodeCharacter{2264}{\ensuremath{\leq}}
\DeclareUnicodeCharacter{2265}{\ensuremath{\geq}}
\DeclareUnicodeCharacter{2282}{\ensuremath{\subset}}
\usepackage{amsmath}
\usepackage{amssymb}
\usepackage{booktabs}
\usepackage{tabularx}
\usepackage{float}
\usepackage{fvextra}
\usepackage{graphicx}
\usepackage{hyperref}
\usepackage[backend=bibtex,natbib=true,style=ext-numeric-comp,giveninits=true,maxbibnames=10,
            sorting=none,language=american]{biblatex}

\AtEveryBibitem{%
  \ifentrytype{article}{%
    \renewbibmacro{in:}{}%

    \renewbibmacro*{issue+date}{%
      \iffieldundef{issue}{}{\printfield{issue}\setunit*{\addcomma\space}}%
      \printdate
    }
  }{}
}
\renewbibmacro*{volume+number+eid}{%
  \printtext[bold]{\printfield{volume}}%
  \iffieldundef{number}{}{\printtext[parens]{\printfield{number}}}%
  \setunit{\bibeidpunct}%
  \printfield{eid}
}

\usepackage{microtype}
\usepackage{tikz}
\usetikzlibrary{positioning}

\newcommand{\F}{\mathcal{F}}
\newcommand{\ep}{p^{(0)}}
\newcommand{\bep}{\mathbf{p}^{(0)}}
\newcommand{\ps}{p^{\star}}
\newcommand{\bps}{\mathbf{p}^{\star}}
\newcommand{\hp}{\hat{p}}
\newcommand{\bhp}{\hat{\mathbf{p}}}
\newcommand{\bp}{\mathbf{p}}
\newcommand{\z}{\mathbf{z}}
\newcommand{\bY}{\mathbf{Y}}
\newcommand{\brier}{\mathsf{Brier}}

\newcommand{\good}[1]{#1}
\newcommand{\bad}[1]{#1}
\newcommand{\mixed}[1]{#1}
\newcommand{\prob}{\mathbb{P}^\star}

\title{Logically Consistent Accurate Forecasts}

\author{
  Kate Yuan\textsuperscript{1,*}
  \quad
  Rohan Alur\textsuperscript{1,2,*}
  \quad
  Manish Raghavan\textsuperscript{2}
  \quad
  Devavrat Shah\textsuperscript{2}
  \\[0.4em]
  \textsuperscript{1}Bridgewater AIA Labs \qquad
  \textsuperscript{2}Massachusetts Institute of Technology
  \\[0.2em]
  {\small \textsuperscript{*}Equal contribution.}
}
\date{}

\begin{document}

\maketitle

\begin{abstract}
  Agentic forecasting systems based on large language models (LLMs) are now competitive with expert human forecasters, but these systems still suffer from a critical limitation: the inability to produce forecasts that are logically \emph{consistent}. For example, forecasts for related events often fail to satisfy the axioms of probability, which in turn implies that related forecasts cannot simultaneously be correct and undermines their utility for decision makers. Enforcing consistency is not trivial --- naive approaches (e.g., projection onto the set of consistent forecasts) can degrade accuracy. We argue that the key to overcoming the apparent tradeoff between consistency and accuracy lies in the reasoning which supports each forecast. We finetune a supervisory language model that leverages this reasoning to jointly reconcile inconsistencies while improving the accuracy of related forecasts. On ForecastBench \cite{karger2024forecastbench}, our method improves the Brier score of a retrieval-augmented GPT-5.1 forecaster by $9.3\%$ while reducing measures of structural inconsistency by $82$--$90\%$. We interrogate our method via a comprehensive series of baselines and ablations, and find that no alternative method matches our gains in consistency and accuracy simultaneously.
  \end{abstract}

\section{Introduction}

Agentic forecasting systems based on large language models (LLMs) have improved rapidly in recent years \cite{karger2024forecastbench,halawi2024approaching,alur2025aia}, but important limitations remain. In this work, we focus on a particular shortcoming: the inability to maintain logical \emph{consistency} across related forecasts. To illustrate this challenge, suppose we ask an LLM-based forecasting system to generate two separate election forecasts: what is the probability a candidate wins by at least $1\%$ of the vote, and what is the probability the same candidate wins by at least $5\%$ of the vote?

We can immediately see that the true probabilities must satisfy $\mathbb{P}(\texttt{wins $\ge 1\%$}) \ge \mathbb{P}(\texttt{wins $\ge 5\%$})$, as the latter necessarily implies the former. However, forecasting systems are typically queried independently on each input, and will commonly violate even elementary probability axioms such as this one. This can be viewed as a special case of modern agentic LLM-based systems' more general failure to maintain a coherent \emph{world model}, even when outputs appear correct both in isolation and on average \cite{vafa2024world, vafa2025inductive, mancoridis2025potemkin}. We discuss additional related work in Appendix~\ref{sec:related-work}.

We argue that this type of inconsistency severely undermines the utility of LLM-based forecasts: forecasts are most valuable insofar as they inform decisions, and it is unclear how to reconcile estimates that are in direct conflict with each other. Unfortunately, while detecting forecast inconsistencies can be straightforward, correcting these inconsistencies without sacrificing the accuracy of forecasts is not --- see the example in Appendix~\ref{sec:intuition} for a concrete illustration. 

In this work we address the challenge of improving logical consistency across questions without trading off accuracy. Our central insight is to examine the LLM forecaster's reasoning when reconciling inconsistent forecasts. We show empirically that this approach can significantly improve both logical consistency and accuracy, overcoming the apparent tradeoff faced by alternative methods.

\begin{figure}[H]
  \centering
  \resizebox{\linewidth}{!}{%
\begin{tikzpicture}[
    font=\scriptsize,
    >=stealth,
    box/.style={draw, rounded corners=2pt, inner sep=2.5pt, minimum height=0.74cm, align=left},
    qbox/.style={box, draw=blue!45, fill=blue!5, text width=3.25cm, align=center},
    fbox/.style={box, draw=orange!60, fill=orange!7, text width=2.6cm, align=center, execute at begin node={\hyphenpenalty=10000\exhyphenpenalty=10000}},
    tag/.style={font=\tiny\itshape, text=black!55},
    hdr/.style={font=\footnotesize\bfseries, align=center},
    sub/.style={font=\tiny, text=black!60, align=center, text width=4.2cm},
    arr/.style={->, thick, black!55},
    bad/.style={text=red!70!black},
  ]
  \def\ra{0}      \def\rb{-0.88} \def\rc{-1.76}
  \def\rd{-2.64}  \def\re{-3.52} \def\rf{-4.40}
  \def\xa{0.97} \def\xb{4.15} \def\xc{8.25} \def\xd{11.6}

  \foreach \x/\h/\s/\w in {
      \xa/{1. Question}/{Given target event\\ to forecast}/1.7cm,
      \xb/{2. Related question cluster}/{Generate structurally\\ related questions}/3.5cm,
      \xc/{3. Upstream forecasts}/{Run the agentic LLM forecaster\\ on each question, producing\\ probabilities and reasoning}/4.3cm,
      \xd/{4. Refined forecasts}/{Jointly reconcile all\\ forecasts given the cluster's\\ probabilities and reasoning}/2.6cm}
  { \node[sub, text width=\w, anchor=north] at (\x,1.72) {\s};
    \node[hdr, anchor=base] at (\x,1.89) {\h}; }

  \node[box, draw=black!50, fill=black!4, text width=1.75cm, align=center, line width=0.8pt]
    (target) at (\xa,{(\ra+\rf)/2})
    {Will candidate X win by at least 1\%?};

  \node[qbox] (q1) at (\xb,\ra) {\textcolor{black!55}{\tiny\itshape nested (stronger)}\\ X wins by ${\ge}\,5\%$?};
  \node[qbox, line width=0.9pt] (q2) at (\xb,\rb) {\textcolor{black!55}{\tiny\itshape original question}\\ \textbf{X wins by ${\ge}\,1\%$?}};
  \node[qbox] (q3) at (\xb,\rc) {\textcolor{black!55}{\tiny\itshape nested (weaker)}\\ X wins (by any margin)?};
  \node[qbox] (q4) at (\xb,\rd) {\textcolor{black!55}{\tiny\itshape complement}\\ X does \emph{not} win by ${\ge}\,1\%$?};
  \node[qbox] (q5) at (\xb,\re) {\textcolor{black!55}{\tiny\itshape partition}\\ X wins by ${<}\,1\%$?};
  \node[qbox] (q6) at (\xb,\rf) {\textcolor{black!55}{\tiny\itshape partition}\\ X loses?};

  \node[fbox] (f1) at (\xc,\ra) {\textbf{50\%} because \ldots};
  \node[fbox] (f2) at (\xc,\rb) {\textbf{40\%} because \ldots};
  \node[fbox] (f3) at (\xc,\rc) {\textbf{62\%} because \ldots};
  \node[fbox] (f4) at (\xc,\rd) {\textbf{72\%} because \ldots};
  \node[fbox] (f5) at (\xc,\re) {\textbf{15\%} because \ldots};
  \node[fbox] (f6) at (\xc,\rf) {\textbf{25\%} because \ldots};
  \foreach \i in {1,...,6} { \draw[arr, thin] (q\i.east) -- (f\i.west); }

  \def\lx{{\xc-0.07}}   %
  \def\ux{{\xc+0.0}}   %
  \def\ax{{(\xc+1.39+\xd-0.5)/2}}   %
  \foreach \k/\lab/\up/\dn in {
      1/nesting:/{$50\% > 40\%$}/{$30\% < 47\%$},
      2/complements:/{sum to $112\%$}/{sum to $99\%$},
      3/partition:/{sum to $80\%$}/{sum to $99\%$}}
  { \pgfmathsetmacro{\yy}{\rf-0.75-0.27*(\k-1)}
    \node[font=\tiny, text=black!60, anchor=base east, inner sep=0pt] at (\lx,\yy) {\lab};
    \node[font=\tiny, bad, anchor=base west, inner sep=0pt] at (\ux,\yy) {\up};
    \node[font=\tiny, text=black!50, anchor=base] at (\ax,\yy) {$\rightarrow$};
    \node[font=\tiny, text=green!35!black, anchor=base, inner sep=0pt] at (\xd,\yy) {\dn}; }

  \node[draw=green!45!black, fill=green!7, rounded corners=3pt,
        minimum width=1.0cm, minimum height={(\ra-\rf)*1cm+0.74cm}]
    (rec) at (\xd,{(\ra+\rf)/2}) {};
  \foreach \y/\p in {\ra/30, \rb/47, \rc/58, \rd/52, \re/11, \rf/41}
    { \node[font=\scriptsize\bfseries] at (\xd,\y) {\p\%}; }

  \draw[arr] (target.east) -- (target.east -| q1.west);
  \draw[arr] ({\xc+1.39},{(\ra+\rf)/2}) -- (rec.west);
\end{tikzpicture}%
}
  \caption{Overview of our method, illustrated with a stylized example.
  Given a target question to forecast, we generate a cluster of related questions with structural relationships to the original. 
  We then run the upstream agentic LLM forecaster on all questions in the cluster, obtaining initial probability forecasts and reasoning traces for each question. 
  Finally, the crux of our method is to jointly reconcile the forecasts within each cluster using the initial probabilities and reasoning traces.}
  \label{fig:method-overview}
\end{figure}
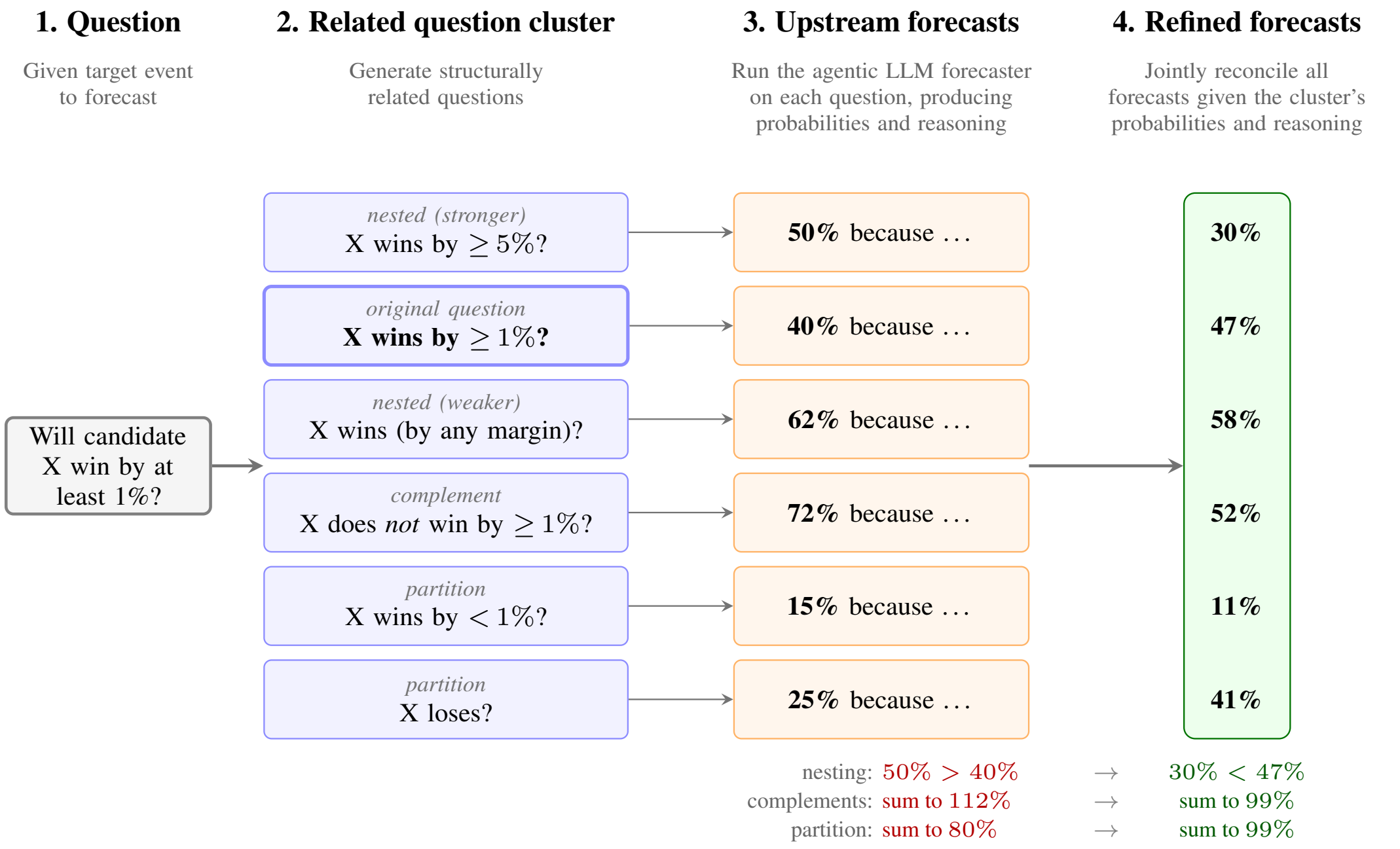

\section{Setup}
\label{sec:setup}

\noindent{\bf Forecasting events.} Let $\Omega$ denote the elementary outcomes of interest, and let $\F$ denote the $\sigma$-algebra over $\Omega$. For example, $\Omega$ may denote the set of possible outcomes for the U.S. Congressional elections, and $\F$ may contain the events these outcomes induce (e.g., that candidate X wins by Y points). Let $\prob$ denote the distribution over outcomes $\Omega$. We are interested in forecasting $n \ge 1$ events, which we denote as $E_i \subset \Omega, E_i \in \F$ for $i \in [n] \equiv \{1, \dots, n\}$. That is, we seek to output the marginal probabilities $\ps_i = \prob(E_i)$ for $i \in [n]$.

\noindent{\bf Estimates and evidence.} For each event $E_i$, let $q_i$ denote the question text describing it. We are provided with a corresponding \emph{probability estimate}  $\ep_i \in [0, 1]$ and a set of accompanying \emph{reasoning} $z_i$. In this work we consider $(\ep_i, z_i)$ that are generated by an upstream agentic forecasting system that has been queried with $q_i$. Thus, $z_i$ denotes the evidence collected by the agentic forecaster (e.g., via open-ended web search) and the reasoning it generates over this evidence for event $E_i$; $\ep_i$ then denotes the upstream forecaster's estimated probability for $E_i$. We let $\bep = [\ep_i : i \in [n]]$ and $\z = [\z_i : i \in [n]]$. We treat these quantities as fixed, exogenously provided inputs throughout.

\noindent{\bf Structural constraints and prior information.} A primary focus of our approach is to leverage relationships between the events of interest, both as inputs to our methodology and for the evaluation thereof. We express these relationships as constraints on the set of allowed probability vectors; for example, the relationship $E_i \subset E_j$ can be expressed as the constraint $\ps_i \le \ps_j$.
We consider three types of structural constraints, which we denote by $\mathcal{C}$, $\mathcal{P}$, and $\mathcal{M}$.\footnote{The complement constraints ($\mathcal{C}$) can be expressed as a special case of the partition constraints ($\mathcal{P}$). We separate them here to match our experimental setup.}
\textbf{Complements} ($\mathcal{C}$): for a pair $(i,j)$ with $E_i = E_j^c$, it must be that $\ps_i + \ps_j = 1$.
\textbf{Partitions} ($\mathcal{P}$): more generally, for $J \subseteq [n]$, if $\bigcup_{j \in J} E_j = \Omega$ and $E_j \cap E_{j'} = \emptyset$ for all $j \neq j' \in J$, then $\sum_{j \in J} \ps_j = 1$. That is, if $J$ denotes a partition of the outcome space, the corresponding forecasts for the events in $J$ must sum to $1$.
\textbf{Nesting} ($\mathcal{M}$): for a chain $E_{i_1} \subseteq \cdots \subseteq E_{i_k}$, we have $\ps_{i_1} \leq \cdots \leq \ps_{i_k}$.

\noindent{\bf Goal and evaluation.} Given $(q_i, \ep_i, z_i : i \in [n])$, our goal is to transform the estimates $\bep$ into refined forecasts $\bhp = [\hp_i : i \in [n]]$. Ideally these refined forecasts will be more accurate: $\bhp$ will be closer to $\bps$ than $\bep$ is. The primary complication, as in all binary forecasting tasks, is that we never observe the true probabilities $\bps$. We do however observe the realized outcomes $Y_i = \mathbf{1}\{E_i \text{ occurs}\}$. By definition, we have $\ps_i = \prob(Y_i = 1) = \prob(E_i)$.

We measure performance via the Brier Score: for $\bp \in [0,1]^n$ and $\bY \in \{0,1\}^n$, $\brier(\bp, \bY) = \frac{1}{n} \sum_{i \in [n]} (p_i - Y_i)^2$. In addition to accuracy, we also evaluate the degree to which the forecasts $\bhp$ conform to the structural constraints above. Specifically, we compute the {\em magnitude of constraint violations} with respect to  $\mathcal{C}, \mathcal{P}$, and $\mathcal{M}$. The precise evaluation metrics are defined in Appendix~\ref{app:loss-function-specifics}.

\section{Method: imposing consistency while improving accuracy}
\label{sec:method}

Our method consists of three stages, as shown in Figure~\ref{fig:method-overview}. First, for each event of interest $E_i$, we produce a set of related events via \emph{synthetic clustering}, described in Section~\ref{sec:synthetic-clustering}. Each resulting cluster consists of the target $E_i$ along with events that are structurally related to $E_i$ via the constraint families $\mathcal{C}$, $\mathcal{P}$ and $\mathcal{M}$. Next, we use an upstream LLM forecaster to generate an initial probability estimate and associated reasoning for each event in the cluster. Finally, we train a small language model to examine these estimates and reasoning jointly and produce an updated forecast $\hp_i$ for each event, described in Section~\ref{sec:learning-process}. Note that this final step is the crux of our method and is agnostic to the particular clustering procedure or upstream LLM forecaster used.

\subsection{Synthetic clustering}
\label{sec:synthetic-clustering}

Each input question $q_i$ is assigned to a cluster consisting of (a) itself and (b) synthetically generated, structurally related variants of itself. For each input question $q_i$, we prompt Claude Opus 4.7 \cite{anthropic2026opus47} to generate related questions corresponding to the constraint families $\mathcal{C}$, $\mathcal{P}$ and $\mathcal{M}$. In particular, for each $q_i$ we generate (a) $\bar{q}_i$ which describes the complement of $q_i$, (b) siblings $q_i^1, \dots, q_i^k$ which, together with $q_i$, form a mutually exclusive and collectively exhaustive partition of the outcome space (i.e., exactly one of the events $q_i, q_i^1, \dots, q_i^k$ must occur),\footnote{Complement questions in (a) are a special case of event-space partitions, so we require $k \geq 2$ in (b).} and (c) a nested sequence of events $\tilde{q}_i^1, \dots, \tilde{q}_i^{\ell}$ which, together with $q_i$, form a chain of increasingly general events (i.e., each event implies the next, so that their probabilities must be monotonically nondecreasing). Appendix~\ref{app:synthetic-generation} contains further details including our validation of the correctness of these clusters. 

Let the set $J_i$ index the synthetic questions associated with $q_i$. Applying the augmentation across all questions yields the index set $I= [n] \cup \bigcup_{i=1}^{n} J_i$. We define $C_i=\{i\}\cup J_i$ as the index set for the cluster corresponding to $q_i$. The set of clusters $\{C_i\}_{i=1}^n$ partitions the augmented index set $I$.

\subsection{Learning algorithm}
\label{sec:learning-process}

Our learning algorithm maps the augmented inputs $\{(q_j,z_j,\ep_j)\}_{j\in I}$ to a set of refined forecasts $\{\hp_j\}_{j\in I}$. For each $j\in I$, let $C(j)$ denote the cluster containing $j$.

To generate a refined forecast for $q_j$, we construct a prompt containing the text, reasoning and estimates for every question in the corresponding cluster $C(j)$, and we set $\hp_j$ to the probability assigned to the ``yes'' token after renormalizing the model's distribution over the ``yes’’ and ``no’’ tokens. Additional details are given in Appendix~\ref{app:learning-process}.

\textbf{Loss function.}
We refine forecasts by finetuning Qwen3.6-35B-A3B \cite{qwen202635b} to minimize an approximately equal weighted sum of three terms -- Brier score, logical consistency penalties, and squared deviation from the upstream forecasts (a regularization term).
The consistency penalty uses squared violations of complement and partition constraints, and hinge violations of monotonicity constraints. 

We do \emph{not} directly observe outcomes for the synthetic questions (although some of these outcomes can be deduced). Thus, Brier score is computed over the original questions only, while consistency and proximity terms are computed over both original and synthetic questions. Appendix~\ref{app:learning-process} contains full details, including exact definitions and hyperparameter settings.

\section{Empirical results}
\label{sec:headline-results}

\textbf{Data.}
\label{sec:data}
We source data from ForecastBench \cite{karger2024forecastbench}, a popular forecasting benchmark of binary questions drawn from prediction markets and well-maintained datasets. 

We construct our training set as the union of unique events released by ForecastBench from 03/02/2025 through 02/15/2026. From this set, we keep only those that resolve strictly before 03/01/2026, and use the questions released from 03/01/2026 through 04/26/2026 as our validation set. This prevents foreknowledge of events in the validation set from bleeding into the training objective. Similarly, we only keep validation set questions which resolve strictly before 05/10/2026, and use questions released from 05/10/2026 through 07/05/2026 as our canonical test set. Importantly, this test set consists only of questions published after the release (and therefore beyond the training cutoff) of Qwen3.6-35B-A3B \cite{qwen202635b}.

Following the evaluation methodology for ForecastBench \cite{karger2024forecastbench}, we compute each evaluation metric separately for two groups, the market sources and dataset sources, and then report their equal-weighted average. Our test set has 243 market questions and 1,344 dataset questions. For efficiency, our training and validation sets sample a random subset of their respective eligible dataset questions, such that the number of dataset questions in each is equal to the number of market questions. See Appendix~\ref{app:forecasting-data} for further details.

\textbf{Upstream forecaster.} We use a retrieval-augmented GPT-5.1~\cite{openai2025gpt51} forecaster. All of our training, validation, and test questions are released after the knowledge cutoff date of GPT-5.1 \cite{openai2025gpt51apidoc}. We use Perplexity as a search provider, and use their API to filter retrieved news sources to those published before the release date of each question \cite{perplexityDateFilters}. Appendix~\ref{app:upstream-forecaster} describes additional details.

\textbf{Results.}
We summarize our primary results in Table~\ref{tab:synthetic-test-results} below. Our method improves the Brier score and all forms of consistency measured on the test set across three independent seeds. The improvement in every metric is significant at the $5\%$ level, detailed in Appendix~\ref{app:significance-testing}. We report results for each seed separately in Appendix~\ref{app:per-seed-results}, and evaluation metric definitions are in Appendix~\ref{app:loss-function-specifics}.

\begin{table}[H]
  \caption{Our method's headline results. \emph{Raw} denotes the upstream forecaster. \emph{Ours} reports the mean $\pm$ sample standard deviation of our method across three seeds. Lower is better for every metric. $\% \Delta$ reports the relative percentage change of \emph{Ours} with respect to \emph{Raw}.}
  \label{tab:synthetic-test-results}
  \centering
    \begin{tabularx}{\textwidth}{@{}l r
      >{\hsize=1.1\hsize\centering\arraybackslash}X
      >{\hsize=.9\hsize\centering\arraybackslash}X@{}}
      \toprule
      Metric                            & Raw      & Ours (mean $\pm$ std)  & $\% \Delta$ (mean $\pm$ std) \\
      \midrule
      Brier score                       & 0.151657 & $0.137519 \pm 0.000900$ & \good{$-9.32\% \pm 0.59\%$}          \\
      Negation squared error            & 0.046831 & $0.005194 \pm 0.000662$ & \good{$-88.91\% \pm 1.41\%$}         \\
      Multi-way partition squared error & 0.223655 & $0.021966 \pm 0.001937$ & \good{$-90.18\% \pm 0.87\%$}         \\
      Monotonicity $L_1$ violation      & 0.018565 & $0.003246 \pm 0.000219$ & \good{$-82.52\% \pm 1.18\%$}         \\
      \bottomrule
    \end{tabularx}
\end{table}

\begin{table}[H]
  \caption{Comparison of our method with baseline methods, showing the relative change ($\% \Delta$) in Brier score and consistency violations with respect to \emph{Raw} for each method. Negative values are better, and the best value in each column is in bold.}
  \label{tab:brier-summary-baselines}
  \centering
    \begin{tabularx}{\textwidth}{@{}
      >{\hsize=1.12\hsize\raggedright\arraybackslash}X
      >{\hsize=.53\hsize\centering\arraybackslash}X
      >{\hsize=1.20\hsize\centering\arraybackslash}X
      >{\hsize=1.15\hsize\centering\arraybackslash}X@{}}
      \toprule
      Method                    & \% $\Delta$ Brier & \% $\Delta$ Pooled partition viol. & \% $\Delta$ Monotonicity viol. \\
      \midrule
      \textbf{Ours}            & \good{$\mathbf{-9.32\%}$}  & \good{${-86.12\%}$}           & \good{${-82.52\%}$}              \\
      Ensemble over 10x Runs   & \good{$-6.57\%$}  & \good{$-41.54\%$}           & \good{$-53.45\%$}              \\
      James--Stein shrinkage    & \good{$-0.88\%$}  & \good{$-2.83\%$}            & \good{$-23.65\%$}              \\
      Platt scaling             & \mixed{$+1.68\%$} & \mixed{$+27.80\%$}          & \mixed{$+5.31\%$}              \\
      Consistent-set projection & \good{$-2.79\%$}  & \good{$\mathbf{-100.0\%}$}           & \good{$\mathbf{-100.0\%}$}              \\
      Zero-Shot LLM             & \mixed{$+3.98\%$} & \good{$\mathbf{-100.0\%}$}           & \good{$\mathbf{-100.0\%}$}              \\
      \bottomrule
    \end{tabularx}
\end{table}

\textbf{Baselines and ablations.}
We conduct a series of baselines and ablations. Our method is statistically better in accuracy, consistency, or both metrics, as compared with every alternative method tested. Table~\ref{tab:brier-summary-baselines} summarizes performance across baseline methods. For example, since we are effectively using additional rollouts of the upstream forecaster when considering related questions' forecasts, we compare against a simple ensemble over ten independent rollouts for the same event.

Table~\ref{tab:brier-summary-ablations} in Appendix~\ref{app:ablation-summary} provides the analogous summary of ablation methods -- a notable ablation is the removal of forecaster reasoning traces from our method, where the Brier change shifts significantly from $-9.32\%$ to $-2.67\%$.
Appendix~\ref{app:baseline-ablation-summary} describes each baseline and ablation, and Appendix~\ref{app:significance-testing} describes significance testing.

\section{Discussion and limitations}
\label{sec:discussion}
In this work, we show that forecasts from LLM-based agentic systems can be post-processed to improve logical consistency without sacrificing accuracy. 
Our key insight is to leverage the forecaster’s own reasoning to reconcile inconsistent forecasts. 
However, our approach has several limitations. The first is scale: our method relies on a relatively small open-source language model, and practical constraints --- both cost and runtime --- preclude us from scaling our method to impose consistency across large numbers of related forecasts. Relatedly, our methodology also assumes that relationships between events are provided exogenously rather than learned directly from the available data, and only handles three types of structural constraints.

These limitations indicate multiple avenues for future research. First, a natural extension of our work would consider a broader set of possible relationships, including ``soft'' constraints on the relationships between forecasts. A second promising direction is to train an auxiliary model to perform this labeling directly, indicating which events should be related and how. Finally, scaling our approach to larger sets of related events --- whether via simply scaling the effective inference compute, or via algorithmic improvements to allow for e.g., iterative refinement of forecasts or inference-time retrieval over related forecasts --- would significantly broaden the applicability of our methods.

\printbibliography

\appendix

\section*{Appendix}

\section{Related work}
\label{sec:related-work}

The present work builds on a rich literature in \emph{judgmental forecasting}, which studies the application of human judgment to forecasting tasks \cite{lawrence2006judgmental, tetlock2005expert, tetlock2015superforecasting}. We focus in particular on binary forecasting, in which the forecaster is presented with a hypothetical future event and tasked with estimating the probability that this event will occur. Judgmental forecasting is often discussed in contrast to statistical forecasting; while the two can often play complementary roles, judgmental forecasting is particularly valuable in domains which lack sufficient data to build statistical models, or in which the available data does not fully capture the relevant dynamics \cite{fildes2002state, lawrence2006judgmental}.

Recent advancements in general-purpose large language models (LLMs) have heralded a significant uptick in work which seeks to \emph{automate} judgmental forecasting. Most modern work on LLM-based forecasters can be traced to the Autocast benchmark \cite{zou2022forecasting}, which found that early language models like T5 \cite{raffel2019t5} and UnifiedQA-v2 \cite{khashabi2022qa} dramatically underperform aggregated human forecasts. Follow-up work similarly found that OpenAI's GPT-4 model \cite{openai2023gpt4} underperforms aggregated human forecasts \cite{schoenegger2023llm, Abolghasemi2023judgmental}. However, later work by \citet{halawi2024approaching} showed that developing an appropriate harness around GPT-4 (or other similarly capable models, like the Claude-2 series \cite{anthropic2023claude2}) could approach -- though not exceed -- the performance of the human crowd.

There are now a wide variety of benchmarks for evaluating the performance of LLM-based forecasters, including the aforementioned Autocast \cite{zou2022forecasting}, the subsequent Autocast++ benchmark \cite{yan2023autocast}, ProphetArena \cite{yang2025prophet}, ForecastBench \cite{karger2024forecastbench}, FutureX \cite{zeng2025future}, the Metaculus Forecasting series \cite{metaculus2025benchmark}, and others. These benchmarks vary significantly in their methodology --- most notably, whether they are static or dynamically update over time, and whether each forecasting question includes the necessary context (e.g., snippets from relevant news articles) or whether the forecaster is charged with retrieving this context itself. In this work we focus on ForecastBench \cite{karger2024forecastbench}, a dynamically updating benchmark which includes only minimal context for each event.

Our work builds directly on a complementary literature focused on methodology for LLM-based forecasters. The first line of work focuses on \emph{retrieval}, or the process by which a forecasting agent gathers relevant context to inform its forecast. \citet{hsieh2024reasoning} extend the static retrieval pipeline of \citet{halawi2024approaching} to allow forecasting agents to dynamically execute their own web search via tool calling; \citet{han2026future} further develop a live reinforcement learning environment for training LLM-based agents to perform open-ended research before producing forecasts. A separate line of work, inspired by earlier work on human superforecasters \cite{tetlock2015superforecasting}, considers ensembling or deliberation among multiple language models to produce more accurate forecasts \cite{schoenegger2024wisdom, li2026diverse}. Finally, \citet{turtel2025llms, turtel2025outcome, scott2026forecasting} propose techniques for directly training language models to output more accurate (and better calibrated) forecasts after being provided with the appropriate context.

Finally, the present work is perhaps most closely related to those which move beyond evaluating forecast accuracy to instead study the \emph{consistency} of language model forecasts. \citet{fluri2023eval} and \citet{pratt2024forecasting} show that early language models failed to produce self-consistent forecasts; for example, by forecasting that sports records will evolve non-monotonically over time, or that language model forecasts for an event and its negation fail to sum to $1$. \citet{paleka2024consistency} propose the use of such consistency checks as an evaluation measure in its own right, and show that consistency correlates with forecast accuracy. Critically, consistency across forecasts can be evaluated instantaneously, while the accuracy of contemporaneous forecasts can only be evaluated later, once the outcomes for the corresponding events are known. The primary focus of our work is to directly impose consistency on related forecasts without degrading their accuracy.

\section{A toy example}
\label{sec:intuition}

We work through a simple toy example to illustrate both the challenges and opportunities entailed by reasoning directly about the relationships between events.

Recall our running example, in which we seek to refine forecasts related to U.S. election outcomes. Let $E_1$ be the event $\{\texttt{candidate X wins by at least $1\%$}\}$ and $E_2$ be $\{\texttt{candidate X wins by at least $5\%$}\}$. Suppose the upstream forecaster gives forecasts of $\ep_1 = 0.40$ and $\ep_2 = 0.50$, respectively.

Even in this highly stylized setting, we can immediately infer that at least one of these forecasts is incorrect. In particular, because $E_2$ necessarily implies $E_1$, the true probabilities should satisfy $\ps_2 \le \ps_1$, but the corresponding estimates violate this constraint. This kind of incoherence is the rule rather than the exception: LLM-based forecasters are queried independently for each event, and have no mechanism for enforcing even elementary forms of consistency across related forecasts. As discussed in Appendix~\ref{sec:related-work}, this can be viewed as an instance of language models' more general failure to form coherent internal models of the world.

While detecting this kind of inconsistency is straightforward, it is far less clear what to do about it. We could impose the constraint by setting $\hp_1 \leftarrow \ep_2$, but this arbitrarily privileges $\ep_2$ over $\ep_1$. Why not set $\hp_2 \leftarrow \ep_1$ instead? More generally, the feasible set $\{ \hp_1, \hp_2 \mid \hp_2 \le \hp_1, 0 \le \hp_1, \hp_2 \le 1 \}$ only weakly constrains the set of allowed forecasts. In the interest of fairness, we could perform a Euclidean projection onto this set, setting $\bhp = \arg\min_{p_2 \le p_1} \|\bp - \bep\|_2 = (0.45, 0.45)$. However, whether this is an improvement depends entirely on the unobserved value of $\bps$. If $\bps = (0.50, 0.40)$, we have improved both forecasts; if $\bps = (0.40, 0.30)$, then $\ep_1$ was correct and we have needlessly corrupted it. The same will be true of any projection method which is agnostic as to the relative quality of $\ep_1$ and $\ep_2$.

This example reveals two things. First, consistency is often desirable in its own right: forecasts are most valuable as inputs to decision making, and mutually contradictory numbers offer little clarity. Second, even if consistency is a secondary concern, examining both forecasts at once reveals an incoherence that we could not have discovered by examining either one in isolation. This in turn offers an opportunity to improve the \emph{accuracy} of our forecasts. However, as this example also reveals, naive projections run the risk of simply ``stealing'' strength (rather than sharing it) across related forecasts.

We argue that the missing ingredient is precisely the one which distinguishes LLM-based forecasts from other forms of statistical forecasting: the evidence and reasoning that accompany each estimate. The two forecasts will in general reflect different, possibly competing perspectives. For example, the reasoning for $\ep_1$ may focus on nomination procedures and intra-party coalition management, while the reasoning for $\ep_2$ may focus on macroeconomic conditions and general election polling. By reasoning over the union of this evidence, we can in principle reconcile it into a single, coherent view of the world. We develop such a method in this paper.

\section{Learning process}
\label{app:learning-process}

Below we describe additional learning process details, building on Section~\ref{sec:learning-process}. 

\subsection{Extracting forecasts}
Instead of directly prompting the model to return a probability, we instruct the model to output ``yes'' if the event described in
$q_j$ is going to occur and ``no'' otherwise. 

We then obtain the refined forecast

\[\hp_j  = \frac{P_\theta(\text{``Yes''}\mid\mathrm{prompt}_j)}{P_\theta(\text{``Yes''}\mid\mathrm{prompt}_j) +P_\theta(\text{``No''}\mid\mathrm{prompt}_j)}
\]

where $\theta$ denotes the model parameters. We use this formulation, rather than
$P_\theta(\text{``Yes''}\mid\mathrm{prompt}_j)$ directly, to exclude the
(typically negligible) probability mass the model assigns to the rest of the
token vocabulary.

\subsection{Metric Definitions and Loss function}
\label{app:loss-function-specifics}

Let $\mathcal{B}\subseteq I$ denote the set of indices in a batch of training
examples, and let $N_{\mathcal{B}}=|\mathcal{B}|$ denote the size of the
batch. We batch by full clusters to ensure that all intra-cluster relations
are present in the same forward pass. Importantly, we do not directly observe
outcomes for the synthetic questions. Thus, the synthetic questions enter the
training objective only through the consistency and proximity terms, which we
detail below. We evaluate the Brier score over the original set of questions
$q_1,\ldots,q_n$.

Let
\[
  \mathcal{R}=\{i\in\mathcal{B}: \exists Y_i\in\{0,1\} ~\mbox{associated with}~i\}
\]
be the subset of events with observed outcomes, and let
$N_{\mathcal{R}}=|\mathcal{R}|$ be the number of such events. Let $\theta$
denote the model parameters and let $\hp_i = \hp_i(\theta)$ denote the refined probability 
the model, parameterized by $\theta$, assigns to event $i$. The loss function combines three terms.

The first term is the Brier score, which evaluates the accuracy of the
forecasts:
\[
  \mathcal{L}_{\mathrm{brier}}(\theta;\mathcal{B})
  =\frac{1}{N_{\mathcal{R}}}
    \sum_{i\in\mathcal{R}}(\hp_i-Y_i)^2.
\]

The second term is the proximity term, which regularizes the refined forecasts
towards the upstream estimates:
\[
  \mathcal{L}_{\mathrm{prox}}(\theta;\mathcal{B})
  =\frac{1}{N_{\mathcal{B}}}
    \sum_{i\in\mathcal{B}}(\hp_i-\ep_i)^2.
\]

The third term is the consistency term, which penalizes violations of the
constraint families $\mathcal{C}$, $\mathcal{P}$, and $\mathcal{M}$. Let
$\mathcal{C}_{\mathcal{B}}$, $\mathcal{P}_{\mathcal{B}}$, and
$\mathcal{M}_{\mathcal{B}}$ denote the subsets of the corresponding constraint
families whose members all appear in the batch $\mathcal{B}$. For example,
\[
  \mathcal{C}_{\mathcal{B}}
  =\{(a,b)\in\mathcal{C}:a,b\in\mathcal{B}\}.
\]
Each constraint family contributes the following terms:
\[
  \begin{aligned}
    \ell_{\mathcal{C}}(a,b)
      &= (\hp_a+\hp_b-1)^2,
      && (a,b)\in\mathcal{C}_{\mathcal{B}},\\
    \ell_{\mathcal{P}}(S)
      &= \left(\sum_{i\in S}\hp_i-1\right)^2,
      && S\in\mathcal{P}_{\mathcal{B}},\\
    \ell_{\mathcal{M}}(m)
      &= \frac{1}{k_m-1}\sum_{r=1}^{k_m-1}
         [\hp_{i_r}-\hp_{i_{r+1}}]_+,
      && m=(i_1,\ldots,i_{k_m})\in\mathcal{M}_{\mathcal{B}},
  \end{aligned}
\]
where complement and partition constraints penalize squared deviations from
the constraint set, while monotone chains use a hinge loss
($[\cdot]_+=\max\{\cdot,0\}$). The monotone hinge loss is averaged over the
$k_m-1$ adjacent members of each chain so that its magnitude is invariant to
chain length.

We weight each term by the corresponding hyperparameters
$\alpha_{\mathcal{C}},\alpha_{\mathcal{P}},\alpha_{\mathcal{M}}\geq 0$ and
define
\[
  N_{\mathrm{cons}}
  =|\mathcal{C}_{\mathcal{B}}|
   +|\mathcal{P}_{\mathcal{B}}|
   +|\mathcal{M}_{\mathcal{B}}|.
\]
This gives the consistency loss
\[
  \mathcal{L}_{\mathrm{cons}}(\theta;\mathcal{B})
  =\frac{1}{N_{\mathrm{cons}}}\left(
    \alpha_{\mathcal{C}}
      \sum_{(a,b)\in\mathcal{C}_{\mathcal{B}}}\ell_{\mathcal{C}}(a,b)
    +\alpha_{\mathcal{P}}
      \sum_{S\in\mathcal{P}_{\mathcal{B}}}\ell_{\mathcal{P}}(S)
    +\alpha_{\mathcal{M}}
      \sum_{m\in\mathcal{M}_{\mathcal{B}}}\ell_{\mathcal{M}}(m)
  \right).
\]
We adopt the convention that
$\mathcal{L}_{\mathrm{cons}}(\theta;\mathcal{B})=0$ when
$N_{\mathrm{cons}}=0$.

Finally, the training loss is a convex combination of the preceding three
terms, with weights
$w_{\mathrm{brier}},w_{\mathrm{prox}},w_{\mathrm{cons}}\geq 0$ satisfying
$w_{\mathrm{brier}}+w_{\mathrm{prox}}+w_{\mathrm{cons}}=1$:
\[
  \mathcal{L}(\theta;\mathcal{B})
  =w_{\mathrm{brier}}\mathcal{L}_{\mathrm{brier}}(\theta;\mathcal{B})
   +w_{\mathrm{prox}}\mathcal{L}_{\mathrm{prox}}(\theta;\mathcal{B})
   +w_{\mathrm{cons}}\mathcal{L}_{\mathrm{cons}}(\theta;\mathcal{B}).
\]

\subsection{Experiment details and hyperparameters}

\begin{itemize}
  \item \textbf{Base model:} \texttt{Qwen3.6-35B-A3B}.
  \item \textbf{Adapter:} LoRA with rank $32$.
  \item \textbf{Prompt rendering:} Qwen chat template with thinking
  disabled.
  \item \textbf{Optimizer:} Adam with a constant learning rate of $10^{-5}$.
  \item \textbf{Epochs:} $2$.
  \item \textbf{Batch Size:} Target batch size of $16$ events. We greedily pack
        complete clusters into each batch, so the realized size of a batch
        may be smaller than $16$ (or larger in the case of a single oversized cluster).
  \item \textbf{Loss weights:}
        $(w_{\mathrm{brier}},w_{\mathrm{prox}},w_{\mathrm{cons}})
          =(0.33,0.34,0.33)$.
  \item \textbf{Consistency scale weights:}
        $(\alpha_{\mathcal C},\alpha_{\mathcal P},
          \alpha_{\mathcal M})=(1,1,5)$ for complement,
        multi-way partition, and monotonicity constraints, respectively.
  \item \textbf{Checkpoint selection:} We evaluate on the validation set every 20 optimization steps. As described in Section~\ref{sec:data}, we follow ForecastBench's methodology of separately computing evaluation metrics on the two source types (market and dataset sources) and then computing their equal-weighted average. Thus, for each checkpoint, we compute the validation loss separately for market and dataset question clusters. Let $\Theta$ denote the set of evaluated checkpoints, and let $\mathcal{V}_{\mathrm{market}}$ and $\mathcal{V}_{\mathrm{dataset}}$ denote the full set of validation questions in market and dataset clusters respectively. We select the checkpoint used for test-set evaluation as
  \[
  \theta^\star \in \arg\min_{\theta\in\Theta}
  \left[
  \tfrac{1}{2}\mathcal{L}(\theta;\mathcal{V}_{\mathrm{market}})
  +\tfrac{1}{2}\mathcal{L}(\theta;\mathcal{V}_{\mathrm{dataset}})
  \right].
  \]
  \item \textbf{Random seed:} $31, 32, 33$, used for adapter initialization and training
        order shuffling.
\end{itemize}

\section{Synthetic question generation}
\label{app:synthetic-generation}

Here we expand on synthetic question generation as described in Section~\ref{sec:synthetic-clustering}. Recall that $J_i$ indexes the synthetic questions associated with $q_i$. We can partition $J_i$ into subsets which correspond to the constraint families $\mathcal{C}$, $\mathcal{P}$, and $\mathcal{M}$ as follows:
\[
  J_i = J_i^{\mathcal{C}} \cup J_i^{\mathcal{P}} \cup J_i^{\mathcal{M}},
\]
where $J_i^{\mathcal{C}}$ indexes the complement $\bar{q}_i$, $J_i^{\mathcal{P}}$ the siblings $q_i^1,\dots,q_i^k$, and $J_i^{\mathcal{M}}$ the nested chain $\tilde{q}_i^1,\dots,\tilde{q}_i^\ell$.

These subsets correspond directly to the corresponding constraint families: $J_i^{\mathcal{C}}$ contributes the complement pair $(i,j) \in \mathcal{C}$, $J_i^{\mathcal{P}}$ gives the partition constraint $\{i\} \cup J_i^{\mathcal{P}} \in \mathcal{P}$, and $J_i^{\mathcal{M}}$ gives the monotone constraints on $\{i\} \cup J_i^{\mathcal{M}}$ in $\mathcal{M}$.

We validate the correctness of these synthetically generated questions with an LLM judge, retaining only those the judge deems valid. To confirm the judge's accuracy, we randomly sample $50$ judge-approved clusters for each type of relationship and manually review them.

This review discovered no false negatives --- events which did not have the desired relationship to the original question but were not flagged by the judge. We can thus upper bound the false negative rate for the (hypothetical) superpopulation at $5.82\%$ for questions corresponding to each constraint set. We obtain this value simply by solving $(1-p)^{50} = 0.05$, which calculates the maximum false negative rate consistent with observing $0$ false negatives over $50$ i.i.d. samples. Alternatively, we can pool all $150$ reviewed cases to bound the false negative rate at $1.98\%$ over all synthetic questions (under the assumption that questions corresponding to the three constraint types are equally represented in our hypothetical superpopulation). We conclude from this exercise that synthetic clustering behaves as desired. 

\begin{table}[H]
    \centering
    \caption{LLM-judge filtering and manual-audit summary for synthetic relationship families in the test set. The final column reports the one-sided $95\%$ confidence bound on the judge's false-pass rate.}
    \label{tab:synthetic-judge-audit}
    \begin{tabularx}{\textwidth}{@{}l
        >{\hsize=.80\hsize\centering\arraybackslash}X
        >{\hsize=1.25\hsize\centering\arraybackslash}X
        >{\hsize=.75\hsize\centering\arraybackslash}X
        >{\hsize=1.00\hsize\centering\arraybackslash}X
        >{\hsize=1.20\hsize\centering\arraybackslash}X@{}}
        \toprule
        Synthetic family & Candidate groups & Judge-approved groups & Groups audited & Audit false passes & $95\%$ judge false-pass rate \\
        \midrule
        Negation             & 1,581 & 1,580 & 50 & 0 & $5.82\%$ \\
        Monotonic ladder     & 1,587 & 1,562 & 50 & 0 & $5.82\%$ \\
        Multi-way partition & 404 & 369 & 50 & 0 & $5.82\%$ \\
        \bottomrule
    \end{tabularx}
\end{table}

\section{Forecasting data}
\label{app:forecasting-data}

In this section we elaborate on our description in Section~\ref{sec:headline-results} about our data sourced from ForecastBench \cite{karger2024forecastbench}, a popular benchmark for evaluating the performance of LLM-based forecasters. As discussed in Appendix~\ref{sec:related-work}, ForecastBench is a dynamic benchmark which regularly publishes new questions about future events. This is intended in part to enable evaluation after the model's pretraining ``knowledge cutoff'', which otherwise contaminates retrospective back testing \cite{sarkar2024lookahead}.

Each question in the dataset prompts the model to forecast whether a particular event will occur by a particular date (the ``resolution date''). Questions span a wide variety of topics, including sports, economics, elections and public policy, technology, and others. They are sourced from a combination of trusted, well-maintained datasets (e.g., the FRED economic database, Yahoo! Finance) and public prediction market platforms (e.g., Manifold Markets and Polymarket). The former are deemed dataset questions and the latter deemed market questions. Performance metrics are measured as a equal-weighted average over the market subset and the dataset subset of questions. We defer to \cite{karger2024forecastbench} for a full treatment of the benchmark construction.

We describe the specific release dates and cutoff criteria we use for each of our training, validation, and test sets in Section~\ref{sec:headline-results}. The same underlying question can be contained in multiple releases, in essence asking the same question at slightly different forecasting dates. Dataset questions can additionally contain multiple resolution-date variants of the same underlying question in a single release. In our test set, we deduplicate instances of the former but not the latter. We randomly select one release date to keep for each underlying question, and then keep all of its resolution-date variants in that release. We use random number generator seed 33.

For our training and validation sets, we do additional deduplication and subsampling for efficiency. First, we randomly select only one release date and one resolution date for each underlying question, deduplicating both types of repetitive questions mentioned above. Then, as described in Section~\ref{sec:headline-results}, we construct our training and validation sets to have an equal number of market and dataset source questions. Since dataset questions outnumber market questions in both sets, we randomly sample a subset of eligible dataset questions for each set, stratifying by original source such that source proportions are maintained. For random number generation, we use seeds 31 and 32 for training and validation respectively. 

\section{Upstream forecaster}
\label{app:upstream-forecaster}

In this section we provide more details  on our upstream forecaster, building on Section~\ref{sec:data}.

\textbf{Retrieval.} We give GPT-5.1 access to agentic web search through the Perplexity API \cite{perplexityAPI}. To prevent foreknowledge leakage, we restrict search results to sources published before the release date of each question by setting fields \texttt{search\_before\_date\_filter} and \texttt{last\_updated\_before\_filter} to that date. As an additional foreknowledge prevention measure, we also explicitly disallow prediction market domains using \texttt{search\_domain\_filter}.

\textbf{Prompt.}
We use GPT-5.1~\cite{openai2025gpt51} with the following system prompt:

\begin{quote}
  \begin{Verbatim}[breaklines=true, breakanywhere=false,breaksymbolleft={}, breakautoindent=false, breakindent=0pt]
You are a well-calibrated forecaster. Estimate P(YES) as of the forecast date under the question text and resolution criteria, the latter of which overrides the former if they conflict.

Answer from the perspective of the forecast date, as if it were the present; do not use knowledge of events or outcomes after that date, even when the forecast date is in the past.

Your estimate will be evaluated by its accuracy (e.g., Brier score), rewarding precision and penalizing overconfidence.

Return these structured fields:
- probability: P(YES) as a number from 0 to 1
- reasoning: a brief explanation of the estimate
- factors_for: 1 to 5 concise reasons supporting YES
- factors_against: 1 to 5 concise reasons supporting NO
- key_uncertainties: 1 to 5 concise unresolved drivers
  \end{Verbatim}
\end{quote}

We use the following user prompt template:

\begin{quote}
  \begin{Verbatim}[breaklines=true, breakanywhere=false,breaksymbolleft={}, breakautoindent=false, breakindent=0pt]
Forecast date: <forecast date>
Resolution date: <resolution date>

Question:
<question>

Resolution criteria:
<resolution criteria>

Background information:
<background information>
  \end{Verbatim}
\end{quote}

\section{Detailed results}
\label{app:per-seed-results}
Table~\ref{tab:synthetic-test-results-per-seed} shows our headline results broken down by seed.

\begin{table}[H]
  \centering
  \caption{Per-seed breakdown of our main results shown in Table~\ref{tab:synthetic-test-results}. \emph{Raw} denotes the upstream forecaster and is shared across seeds; lower is better for every metric. Each column reports the metric value with its relative change from Raw in parentheses (negative is better). Brier score is computed over the resolved original questions, whereas consistency metrics are computed over the corresponding augmented clusters. Each metric is an equal-weighted average over the $243$-question market subset and the $1,344$-question dataset subset.}
  \label{tab:synthetic-test-results-per-seed}
    \begin{tabularx}{\textwidth}{@{}
      >{\hsize=1.12\hsize\raggedright\arraybackslash}X
      r
      *{3}{>{\hsize=.96\hsize\centering\arraybackslash}X}@{}}
      \toprule
      Metric                            & Raw      & Seed 31                        & Seed 32                        & Seed 33                        \\
      \midrule
      Brier score                       & 0.1517 & $0.1385$ \good{$(-8.65\%)$}  & $0.1372$ \good{$(-9.56\%)$}  & $0.1369$ \good{$(-9.76\%)$}  \\
      Negation sq.\ error               & 0.0468 & $0.0056$ \good{$(-88.09\%)$} & $0.0044$ \good{$(-90.54\%)$} & $0.0056$ \good{$(-88.09\%)$} \\
      Partition sq.\ error    & 0.2237 & $0.0201$ \good{$(-91.00\%)$} & $0.0240$ \good{$(-89.27\%)$} & $0.0218$ \good{$(-90.27\%)$} \\
      Monotonicity $L_1$ viol.          & 0.0186 & $0.0035$ \good{$(-81.31\%)$} & $0.0030$ \good{$(-83.66\%)$} & $0.0032$ \good{$(-82.59\%)$} \\
      \bottomrule
    \end{tabularx}

\end{table}

We report the three-seed mean
separately for market and dataset questions in Table~\ref{tab:synthetic-test-results-by-source}.

\begin{table}[H]
  \centering
  \caption{Our main results by question source. Each entry reports
  Raw $\rightarrow$ Model, with the relative change from Raw in parentheses.
  Model values are means across the three training seeds. Lower is better for
  every metric.}
  \label{tab:synthetic-test-results-by-source}
  \begin{tabularx}{\textwidth}{@{}
    >{\hsize=.9\hsize\raggedright\arraybackslash}X
    *{2}{>{\hsize=1.05\hsize\centering\arraybackslash}X}@{}}
    \toprule
    Metric & Market ($n=243$) & Dataset ($n=1{,}344$) \\
    \midrule
    Brier score
      & $0.1451 \rightarrow 0.1291$ \good{$(-11.08\%)$}
      & $0.1582 \rightarrow 0.1460$ \good{$(-7.71\%)$} \\
    Negation squared error
      & $0.0773 \rightarrow 0.0078$ \good{$(-89.92\%)$}
      & $0.0164 \rightarrow 0.0026$ \good{$(-84.12\%)$} \\
    Partition squared error
      & $0.1198 \rightarrow 0.0249$ \good{$(-79.19\%)$}
      & $0.3275 \rightarrow 0.0190$ \good{$(-94.20\%)$} \\
    Monotonicity $L_1$ violation
      & $0.0201 \rightarrow 0.0047$ \good{$(-76.65\%)$}
      & $0.0170 \rightarrow 0.0018$ \good{$(-89.43\%)$} \\
    \bottomrule
  \end{tabularx}
\end{table}

Below we also show the Brier change of the complement questions, since their outcomes can be derived. We include a
seed-level breakdown. Averaged across seeds, our method reduces complement
Brier score by $20.05\%$. This improvement is statistically significant under
a paired bootstrap test (two-sided $p=0.0002$ with 10,000 bootstrap replicates). See
Appendix~\ref{app:significance-testing} for uncertainty quantification methodology.

\begin{table}[H]
  \centering
  \caption{Brier score over the complement questions in our test set, computed as an equal-weighted average over the $243$-question market subset and $1,344$-question dataset subset. \emph{Raw} denotes the upstream forecaster and is shared across seeds. $\Delta$ is Model minus Raw, so negative values indicate improvement. The average row is the mean of the per-seed squared losses.}
  \label{tab:complement-brier-per-seed}
  \begin{tabular}{@{}lrrrr@{}}
    \toprule
    Seed & Raw & Ours & $\Delta$ & $\%\Delta$ \\
    \midrule
    Average & 0.176466 & 0.141092 & \good{$-0.035373$} & \good{$-20.05\%$} \\
    31              & 0.176466 & 0.142999 & \good{$-0.033467$} & \good{$-18.96\%$} \\
    32              & 0.176466 & 0.140033 & \good{$-0.036433$} & \good{$-20.65\%$} \\
    33              & 0.176466 & 0.140246 & \good{$-0.036220$} & \good{$-20.53\%$} \\
    \bottomrule
  \end{tabular}
\end{table}

\section{Baselines and ablations}
\label{app:baseline-ablation-summary}

\subsection{Baselines}
\label{app:baseline-summary}
We consider five baseline methods. As Table~\ref{tab:brier-summary-baselines} illustrates, none of the baseline methods match our method in both consistency and accuracy improvements, and Appendix~\ref{app:significance-testing} demonstrates that the gap is statistically significant. Here we provide a brief description of each baseline. More detailed results for each baseline are provided in Appendix~\ref{app:baselines-additional}.

\textbf{Ensemble scaling.} Recall that, for each event $E_i$, the input to our method consists of reasoning $z_i$ along with the probability estimate $\ep_i$ output by an upstream forecaster. We ask whether producing an ensemble over ten independent rollouts for the \emph{same} event is sufficient to reproduce our results. This baseline is motivated by \citet{alur2025aia}, which attributes substantial improvements in accuracy to this form of ensembling. Because our method also requires additional rollouts of the upstream forecaster,\footnote{Across our $1,587$ test clusters, the mean cluster size is $7.23$ questions (with a standard deviation of $1.63$), including the original question.} ensemble scaling tests whether its gains arise merely from a similar form of ``test-time scaling,'' rather than from meaningfully leveraging relationships between different events. 

\textbf{James--Stein shrinkage.} Next, we consider whether our method can be distinguished from a statistical correction which simply regularizes each estimate toward a fixed point, discarding the associated reasoning as well as the nature of the relationships between events. In particular, we consider positive-part James-Stein shrinkage, which we implement by regularizing each initial estimate $\ep_i$ toward the mean of the corresponding cluster $\bar p_c$:

\[
  \hat p_i=(1-\lambda_c)\ep_i+\lambda_c\bar p_c,
  \qquad
  \lambda_c=
  \min\left\{1,\max\left\{0,
  \frac{(k_c-3)\bar\sigma_c^2}
  {\sum_{i\in c}(\ep_i-\bar p_c)^2}
  \right\}\right\},
\]

where $k_c=|c|$ is the cluster size. To estimate $\bar\sigma_c^2$, we use the same ten independent rollouts of each question that we used for the ensemble scaling baseline. Let $p_{ir}$ denote the forecast from rollout $r\in[10]$ for event $E_i$. Recall that $\ep_i=p_{i1}$. Let $\bar p_i=\frac{1}{10}\sum_{r=1}^{10}p_{ir}$. We estimate the variance $\bar\sigma_c^2$ as
\[
  \hat\sigma_i^2
  =\frac{1}{(10-1)}
  \sum_{r=1}^{10}(p_{ir}-\bar p_i)^2,
  \qquad
  \bar\sigma_c^2
  =\frac{1}{k_c}\sum_{i\in c}\hat\sigma_i^2.
\]
We consider shrinking towards zero as opposed to the cluster mean, as well as heteroskedastic and log-odds variants of the James--Stein estimator in Appendix~\ref{app:shrinkage-baselines}.

\textbf{Platt scaling.} Next we consider \emph{Platt Scaling}, which instead \emph{extremizes} forecasts towards either $0$ or $1$. This approach is motivated by the empirical observation that language models tend to hedge binary forecasts towards $0.5$, and are otherwise subject to various forms of miscalibration \cite{alur2025aia,pratt2024forecasting}. Following \cite{alur2025aia}, we set
\[
  \hat p_i=\sigma\!\left(\sqrt{3}\operatorname{logit}(\ep_i)\right).
\]

Like the James-Stein shrinkage baseline, this statistical correction operates only on the estimates $\ep_i$ and discards the rest of the input.

\textbf{Projection onto the consistent set.} We consider a Euclidean projection of the estimates $\ep_i$ onto the set of consistent forecasts. Let $\mathcal{K}\subseteq[0,1]^{|I|}$ denote the set of forecasts which satisfy the constraints in $\mathcal{C}, \mathcal{P}$ and $\mathcal{M}$. We compute
\[
  \hat{\bp}
  =\arg\min_{\bp\in\mathcal{K}}
  \|\bp-\bep\|_2^2.
\]

This projection enforces the constraints via the smallest possible perturbation to the input estimates. As discussed in Appendix~\ref{sec:intuition}, this kind of projection does not privilege one estimate over another; in particular, it does not leverage the reasoning $\z$ to distinguish which estimates might be more accurate than others. We report additional variations of this baseline in Appendix~\ref{app:projection-baselines}.

\textbf{Zero-shot LLM.} Finally, we prompt GPT-5.1 (the same LLM which underpins the upstream forecaster) with each cluster of questions, their initial forecasts, and all the explicit consistency relationships to enforce, and we ask for updated probabilities for all the questions that are reconciled to be consistent. The full prompt is provided in Appendix~\ref{app:zero-shot-llm}.

\subsection{Ablations}
\label{app:ablation-summary}

We now ablate the three primary components of our methodology: providing the model with cross-question context, imposing an explicit consistency objective, and incorporating reasoning traces from the upstream forecaster. These ablation experiments reveal that each of these three ingredients is necessary to achieve our improvements in both accuracy and consistency across forecasts. We also substitute our finetuned model with GPT-5.1 off-the-shelf and find that it underperforms our finetuned model. 
Below we provide a brief description of each ablation. We summarize the results of these ablation experiments in Table~\ref{tab:brier-summary-ablations} below. Additional details for each ablation are provided in Appendix~\ref{app:detailed-ablations}.

\textbf{No cross-question context.} We first consider removing the model's ability to reconcile context across distinct but related questions.
Instead, given a target event $E_i$ and corresponding question $q_i$, we retain the original upstream forecast $(p_i^{(0)}, z_i)$ and generate nine additional independent rollouts $(p_i^{(r)}, z_i^{(r)})$ for $r \in \{1, \ldots, 9\}$ for the \emph{same} event.
Each set of ten forecasts forms a distinct cluster, allowing our model to reconcile the reasoning across multiple upstream forecasts while preventing it from leveraging relationships between distinct events.
The loss function substitutes the cross-question consistency term for the average of the squared penalties on each pairwise difference between predicted probabilities within the cluster. For a clean comparison with our main experiment and to isolate the effect of cross-question context from ensembling (which is a separate baseline discussed in Appendix~\ref{app:baseline-summary}), we compute the updated Brier score using only the updated forecast $\hat{p}_i$ corresponding to the original upstream forecast $p_i^{(0)}$, rather than ensembling the updated forecasts from all ten rollouts. We measure consistency across related questions by similarly generating nine additional rollouts of each synthetic question, running them through this ablation model, and calculating consistency metrics on the $\hp_i$ estimates of each original question and its synthetic related questions.

\textbf{No explicit consistency objective.} Next we consider whether the model can learn to produce mutually coherent forecasts without an explicit supervision signal for consistency. In particular, we remove the consistency term from the loss function while maintaining the overall weight of the regularization term, which gives the following modified training objective:
\[
  \mathcal{L}(\theta;\mathcal{B})
  =
  0.66\,\mathcal{L}_{\mathrm{brier}}(\theta;\mathcal{B})
  +
  0.34\,\mathcal{L}_{\mathrm{prox}}(\theta;\mathcal{B}).
\]

The training and evaluation setup is otherwise as described in Section~\ref{sec:headline-results}. In particular, the model is still presented with estimates and reasoning traces across distinct but related questions, and can in principle learn to produce consistent forecasts as a consequence of maximizing accuracy.

\textbf{No reasoning traces.} Next, we revoke the model's ability to inspect the reasoning produced by the upstream forecaster. In this experiment, we simply remove the trace $z_i$ from each cluster while otherwise retaining the structure of the prompt. The model still retains access to the questions $q_i$ and the corresponding estimates $\ep_i$, and can therefore learn both to identify inconsistent forecasts and to perform statistical corrections. However, it cannot inspect the evidence used to produce each estimate to decide \emph{how} to reconcile inconsistent forecasts.

\textbf{Frontier LLM.} Finally, we consider the use of a frontier language model. We again choose OpenAI's GPT-5.1 \cite{openai2025gpt51} model to remain consistent with the upstream forecaster. We prompt this model identically to our own method, with additional instructions to directly provide a probability estimate. Recall, as described in Section~\ref{sec:learning-process} that our method infers these probabilities from the logits corresponding to the ``YES'' and ``NO'' tokens; this is infeasible for GPT-5.1, which does expose these logits.

\begin{table}[H]
  \caption{The relative improvement of each ablation method with respect to \emph{Raw}; negative values indicate better performance. \textbf{Ours} denotes our method, with metrics averaged over three seeds.}
  \label{tab:brier-summary-ablations}
  \centering
    \begin{tabularx}{\textwidth}{@{}
      >{\hsize=1.65\hsize\raggedright\arraybackslash}X
      >{\hsize=.55\hsize\centering\arraybackslash}X
      >{\hsize=.95\hsize\centering\arraybackslash}X
      >{\hsize=.85\hsize\centering\arraybackslash}X@{}}
      \toprule
      Method                       & \% $\Delta$ Brier & \% $\Delta$ Pooled partition viol. & \% $\Delta$ Monotonicity viol. \\
      \midrule
      \textbf{Ours}                & \good{${-9.32\%}$}           & \good{$\mathbf{-86.12\%}$} & \good{$\mathbf{-82.52\%}$} \\
      Ablation: No consistency loss       & \good{$\mathbf{-9.69\%}$} & \good{$-51.68\%$} & \good{$-37.38\%$} \\
      Ablation: No reasoning traces       & \good{$-2.67\%$}           & \good{$-81.72\%$} & \good{$-79.37\%$} \\
      Ablation: No cross-question context & \good{$-8.88\%$}           & \good{$-57.75\%$} & \good{$-49.78\%$} \\
      Ablation: Frontier GPT-5.1          & \good{$-5.83\%$}           & \good{$-18.26\%$} & \good{$-2.16\%$}  \\
      \bottomrule
    \end{tabularx}
\end{table}

\section{Significance testing}
\label{app:significance-testing}

We compute pairwise bootstrapped p-values between our model and the raw upstream forecasts, as well as between our model and each baseline method, following the methodology of ~\cite{karger2024forecastbench} and ~\cite{alur2025aia}.
This test based on pairwise differences is more informative than a traditional bootstrapped confidence interval around Brier score because in the latter case most of the variation is driven by inter-question variability, and two models' confidence intervals can look quite similar even if one model strictly outperforms the other on every question.

We compute the difference in Brier score between our method and each alternative method for each question in the test set.
We then bootstrap sample with replacement $N=10{,}000$ times, stratifying by market-question subset and dataset-question subset, such that each sample is the same size as the test set and contains $243$ market questions and $1,344$ dataset questions.
We compute the average difference in Brier score for each bootstrap sample to get a distribution of averages (when we compute the average, we separately calculate the market subset average and the dataset subset average, and then equal-weight their combination). 
We recenter this distribution at zero by subtracting average of averages (the average of the bootstrap sample average differences).
Then, we compute the two-sided p-value of our original mean of Brier score differences (equal-weighted between market and dataset sources) on this recentered bootstrap distribution. 

We similarly compute the paired bootstrap p-value between our method and each alternative method for each consistency metric. We use the same bootstrap samples across all metrics -- clusters are sampled as a unit to preserve intra-cluster relationships.

\paragraph{Method vs. raw.}

As Table~\ref{tab:raw-pairwise-pvalues} illustrates, our method is significantly better in all metrics as compared with the raw upstream forecasts.

\begin{table}[H]
  \centering
  \caption{Two-sided paired-bootstrap p-values comparing our method with the
    raw upstream forecasts on the test set. Bold entries are significant at
    the 5\% level; the superscript P indicates that our method has the lower
    metric.}
  \label{tab:raw-pairwise-pvalues}
  \begin{tabular}{@{}cccc@{}}
    \toprule
    Brier & Negation sq. error & Multi-way partition sq. error & Monotonicity $L_1$ violation \\
    \midrule
    $\mathbf{0.0012}^{\mathrm{P}}$ & $\mathbf{0.0002}^{\mathrm{P}}$ & $\mathbf{0.0002}^{\mathrm{P}}$ & $\mathbf{0.0002}^{\mathrm{P}}$ \\
    \bottomrule
  \end{tabular}
\end{table}

\paragraph{Baselines.}

Our method is significantly better on all computed metrics compared with James--Stein
shrinkage and Platt scaling. Our method has significantly lower consistency violations on all three measures as compared with ensemble scaling. For the two methods that impose the constraints exactly, consistent-set projection and
the zero-shot LLM, our method  has significantly lower Brier score. 

\begin{table}[H]
  \centering
  \caption{Two-sided paired-bootstrap p-values comparing our method
    with each baseline on the test set.  Bold entries are significant at the
    5\% level; the superscript indicates whether our method (P) or the
    baseline (B) has the lower metric. }
  \label{tab:baseline-pairwise-pvalues}
  \begin{tabularx}{\textwidth}{@{}l
    >{\hsize=.65\hsize\centering\arraybackslash}X
    >{\hsize=.80\hsize\centering\arraybackslash}X
    >{\hsize=1.30\hsize\centering\arraybackslash}X
    >{\hsize=1.25\hsize\centering\arraybackslash}X@{}}
    \toprule
    Baseline & Brier & Negation sq. error & Multi-way partition sq. error & Monotonicity $L_1$ violation \\
    \midrule
    Ensemble over 10x Runs    & 0.2088 & $\mathbf{0.0002}^{\mathrm{P}}$ & $\mathbf{0.0002}^{\mathrm{P}}$ & $\mathbf{0.0002}^{\mathrm{P}}$ \\
    James--Stein shrinkage    & $\mathbf{0.0010}^{\mathrm{P}}$ & $\mathbf{0.0002}^{\mathrm{P}}$ & $\mathbf{0.0002}^{\mathrm{P}}$ & $\mathbf{0.0002}^{\mathrm{P}}$ \\
    Platt scaling             & $\mathbf{0.0024}^{\mathrm{P}}$ & $\mathbf{0.0002}^{\mathrm{P}}$ & $\mathbf{0.0002}^{\mathrm{P}}$ & $\mathbf{0.0002}^{\mathrm{P}}$ \\
    Consistent-set projection & $\mathbf{0.0014}^{\mathrm{P}}$ & $\mathbf{0.0002}^{\mathrm{B}}$ & $\mathbf{0.0002}^{\mathrm{B}}$ & $\mathbf{0.0002}^{\mathrm{B}}$ \\
    Zero-Shot GPT-5.1         & $\mathbf{0.0006}^{\mathrm{P}}$ & $\mathbf{0.0002}^{\mathrm{B}}$ & $\mathbf{0.0002}^{\mathrm{B}}$ & $\mathbf{0.0002}^{\mathrm{B}}$ \\
    \bottomrule
  \end{tabularx}
\end{table}

\paragraph{Ablations.}

Our method has significantly lower Brier score than the no-reasoning-trace ablation.  It has significantly lower consistency violations than the no-consistency-loss, no-cross-question-context, and the frontier LLM ablation.

\begin{table}[H]
  \centering
  \caption{Two-sided paired-bootstrap p-values comparing our method with each
    ablation on the test set.  Bold entries are significant at the 5\% level;
    the superscript indicates whether our method (P) or the ablation (A) has
    the lower metric. }
  \label{tab:ablation-pairwise-pvalues}
  \begin{tabularx}{\textwidth}{@{}l
    >{\hsize=.65\hsize\centering\arraybackslash}X
    >{\hsize=.80\hsize\centering\arraybackslash}X
    >{\hsize=1.30\hsize\centering\arraybackslash}X
    >{\hsize=1.25\hsize\centering\arraybackslash}X@{}}
    \toprule
    Ablation & Brier & Negation sq. error & Multi-way partition sq. error & Monotonicity $L_1$ violation \\
    \midrule
    No consistency loss          & 0.7039 & $\mathbf{0.0002}^{\mathrm{P}}$ & $\mathbf{0.0002}^{\mathrm{P}}$ & $\mathbf{0.0002}^{\mathrm{P}}$ \\
    No reasoning traces          & $\mathbf{0.0002}^{\mathrm{P}}$ & $0.0644^{\mathrm{P}}$ & $\mathbf{0.0020}^{\mathrm{P}}$ & $\mathbf{0.0200}^{\mathrm{P}}$ \\
    No cross-question context & 0.7763 & $\mathbf{0.0002}^{\mathrm{P}}$ & $\mathbf{0.0002}^{\mathrm{P}}$ & $\mathbf{0.0002}^{\mathrm{P}}$ \\
    Frontier GPT-5.1             & 0.1738 & $\mathbf{0.0002}^{\mathrm{P}}$ & $\mathbf{0.0002}^{\mathrm{P}}$ & $\mathbf{0.0002}^{\mathrm{P}}$ \\
    \bottomrule
  \end{tabularx}
\end{table}

\section{Additional baseline details}
\label{app:baselines-additional}

Here we provide additional details on the baseline methods mentioned in Appendix~\ref{app:baseline-summary}.

\subsection{Ensemble scaling}
We investigate whether the increased number of independent upstream LLM forecaster runs that are now feeding each individual updated forecast is responsible for our model's gains.
As a baseline, we run ten times as many independent upstream forecaster runs for each original question and take their ensemble average. This results in Brier gains and consistency gains, but they are not as strong as our method. See Section~\ref{app:significance-testing} for significance testing. 

\begin{table}[H]
  \centering
  \caption{Arithmetic mean of ten independent GPT-5.1 + Perplexity forecasts on the test set.}
  \label{tab:same-question-more-runs}
  \begin{tabularx}{\textwidth}{@{}l*{4}{>{\centering\arraybackslash}X}@{}}
      \toprule
      Metric                            & Raw      & 10-run mean & $\Delta$            & Percent $\Delta$          \\
      \midrule
      Brier score                       & 0.151657 & 0.141696 & \good{$-0.009961$} & \good{$-6.57\%$}  \\
      Negation squared error            & 0.046831 & 0.011157 & \good{$-0.035674$} & \good{$-76.18\%$} \\
      Multi-way partition squared error & 0.223655 & 0.176326 & \good{$-0.047330$} & \good{$-21.16\%$} \\
      Monotonicity $L_1$ violation      & 0.018565 & 0.008642 & \good{$-0.009923$} & \good{$-53.45\%$} \\
      \bottomrule
  \end{tabularx}
\end{table}

\subsection{Shrinkage}
\label{app:shrinkage-baselines}

We investigate whether our method is merely implementing a form of regularization via shrinkage.
Assuming we are given the same clusters of related questions as described in section~\ref{sec:synthetic-clustering}, we can directly apply James-Stein style shrinkage \cite{james1961estimation} to the forecasts of each question in the cluster.

For each cluster, we shrink each initial forecast $\ep_i$ toward the cluster mean $\bar p_c$:

\[
  \hat p_i = (1-\lambda_c)p_i^{(0)} + \lambda_c\bar p_c.
\]
We use the classical positive-part James-Stein coefficient \cite{baranchik1970family}, which assumes a common variance across questions within each cluster:
\[
  \lambda_c = \min\left\{1, \max\left\{0, \frac{(k_c-3)\bar\sigma_c^2}{\sum_{i\in c}(p_i^{(0)}-\bar p_c)^2}\right\}\right\}.
\]
Here $k_c$ is the number of questions in cluster $c$ and $\bar\sigma_c^2$ is the average sample variance across ten rollouts for each question in the cluster.

We see that shrinkage toward the cluster mean does not meaningfully improve Brier score.

\begin{table}[H]
  \centering
  \caption{Classical positive-part James-Stein shrinkage on synthetic-question clusters.}
  \label{tab:classical-js-shrinkage}
  \begin{tabularx}{\textwidth}{@{}l*{4}{>{\centering\arraybackslash}X}@{}}
      \toprule
      Metric                        & Raw      & Classical JS & $\Delta$           & Percent $\Delta$          \\
      \midrule
      Brier score                       & 0.151657 & 0.150317 & \good{$-0.001339$} & \good{$-0.88\%$}  \\
      Negation squared error            & 0.046831 & 0.039709 & \good{$-0.007122$} & \good{$-15.21\%$} \\
      Multi-way partition squared error & 0.223655 & 0.220809 & \good{$-0.002846$} & \good{$-1.27\%$}  \\
      Monotonicity $L_1$ violation      & 0.018565 & 0.014174 & \good{$-0.004391$} & \good{$-23.65\%$} \\
      \bottomrule
  \end{tabularx}
\end{table}

The properties of our initial probability forecasts don't satisfy homoskedastic assumptions, or assumptions on being unbounded in value.
Below we report heteroskedastic empirical Bayes and log-odds variations of this experiment.

\paragraph{Heteroskedastic empirical Bayes.}
We also test a heteroskedastic empirical Bayes form of the shrinkage estimator \cite{efron1975data}.
Questions can have different levels of uncertainty, and we have empirical estimates of those uncertainty levels from the ten independent runs of each question.
For each cluster, we shrink each initial forecast $p_i^{(0)}$ toward the cluster mean $\bar p_c$:
\[
  \hat p_i = (1-\lambda_i)p_i^{(0)} + \lambda_i \bar p_c,
  \qquad
  \lambda_i = \frac{\hat\sigma_i^2}{\hat\sigma_i^2 + \hat\tau_c^2}.
\]
Here $\hat\sigma_i^2$ is the sample variance across the ten rollouts for question $i$, and $\hat\tau_c^2$ is an estimate of the inter-question dispersion within cluster $c$,
\[
  \hat\tau_c^2 = \max\!\big(0,\; S_c^2 - \overline{\hat\sigma^2}_c\big),
  \qquad
  S_c^2 = \frac{1}{k_c-1}\sum_{i\in c}\big(p_i^{(0)} - \bar p_c\big)^2,
\]
where $S_c^2$ is the observed dispersion of the raw forecasts within cluster $c$ and $\overline{\hat\sigma^2}_c$ is the average intra-question sampling noise in that cluster.

\begin{table}[H]
  \centering
  \caption{Heteroskedastic empirical Bayes shrinkage on synthetic-question clusters.}
  \label{tab:empirical-js-shrinkage}
  \begin{tabularx}{\textwidth}{@{}l*{4}{>{\centering\arraybackslash}X}@{}}
      \toprule
      Metric                        & Raw      & Shrunk   & $\Delta$           & Percent $\Delta$          \\
      \midrule
      Brier score                       & 0.151657 & 0.149164 & \good{$-0.002492$} & \good{$-1.64\%$}  \\
      Negation squared error            & 0.046831 & 0.040904 & \good{$-0.005927$} & \good{$-12.66\%$} \\
      Multi-way partition squared error & 0.223655 & 0.198432 & \good{$-0.025223$} & \good{$-11.28\%$} \\
      Monotonicity $L_1$ violation      & 0.018565 & 0.014030 & \good{$-0.004535$} & \good{$-24.43\%$} \\
      \bottomrule
  \end{tabularx}
\end{table}

\paragraph{Log-odds classical James-Stein.}
The main shrinkage baseline in Table~\ref{tab:classical-js-shrinkage} applies classical positive-part James-Stein shrinkage directly to the initial probabilities $p_i^{(0)}$.
We also try applying the same classical shrinkage rule to the log-odds, $\operatorname{logit}(p_i^{(0)})$, which are unbounded.
\begin{table}[H]
  \centering
  \caption{Log-odds classical positive-part James-Stein shrinkage on synthetic-question clusters.}
  \label{tab:logodds-classical-js-shrinkage}
  \begin{tabularx}{\textwidth}{@{}l*{4}{>{\centering\arraybackslash}X}@{}}
      \toprule
      Metric                        & Raw      & Log-odds Classical JS & $\Delta$           & Percent $\Delta$          \\
      \midrule
      Brier score                       & 0.151657 & 0.155155 & \mixed{$+0.003498$} & \mixed{$+2.31\%$} \\
      Negation squared error            & 0.046831 & 0.078087 & \bad{$+0.031256$}   & \bad{$+66.74\%$}   \\
      Multi-way partition squared error & 0.223655 & 0.188271 & \good{$-0.035384$}  & \good{$-15.82\%$}  \\
      Monotonicity $L_1$ violation      & 0.018565 & 0.010500 & \good{$-0.008065$}  & \good{$-43.44\%$}  \\
      \bottomrule
  \end{tabularx}
\end{table}

The heteroskedastic empirical Bayes and log-odds variants do not meaningfully improve performance, so the shrinkage baseline's failure does not appear to be specific to the probability-scale classical James-Stein form.

\paragraph{Shrinking towards zero.}
Finally, we try shrinking probabilities towards zero instead of the cluster mean. Let $K$ be the total number of
questions in the test set, including synthetic questions. 
We apply one global shrinkage coefficient,
\[
  \hat p_i=(1-\lambda)p_i^{(0)},
  \qquad
  \lambda=
  \min\left\{1,
  \frac{(K-2)\bar\sigma^2}
       {\sum_{i=1}^{K}(p_i^{(0)})^2}
  \right\}.
\]
Here $K=11{,}472$ and $\bar\sigma^2$ is the average sample variance of forecasts across all questions.
Table~\ref{tab:classical-js-zero-target-shrinkage} shows the performance of shrinking towards zero, which yields similar performance to the other forms of shrinkage shown.

\begin{table}[H]
  \centering
  \caption{Global classical positive-part James--Stein shrinkage toward the fixed probability target zero.}
  \label{tab:classical-js-zero-target-shrinkage}
  \begin{tabularx}{\textwidth}{@{}l*{4}{>{\centering\arraybackslash}X}@{}}
      \toprule
      Metric & Raw & Zero-target Classical JS & $\Delta$ & Percent $\Delta$ \\
      \midrule
      Brier score                       & 0.151657 & 0.149828 & \good{$-0.001829$} & \good{$-1.21\%$}  \\
      Negation squared error            & 0.046831 & 0.042051 & \good{$-0.004779$} & \good{$-10.21\%$} \\
      Multi-way partition squared error & 0.223655 & 0.178662 & \good{$-0.044993$} & \good{$-20.12\%$} \\
      Monotonicity $L_1$ violation      & 0.018565 & 0.017647 & \good{$-0.000917$} & \good{$-4.94\%$}  \\
      \bottomrule
  \end{tabularx}
\end{table}

\subsection{Platt scaling}
\label{app:platt-scaling-analysis}

In the opposite direction to shrinkage, there's the question of whether we are simply improving the calibration of predictions by extremizing them, since LLMs tend to hedge towards 0.5 \cite{alur2025aia}.
Following the fixed-parameter convention used in \cite{alur2025aia}, we apply Platt scaling with coefficient $\sqrt{3}$ directly to each initial ensemble mean forecast:
\[
  \hat p_i = \sigma\!\left(\sqrt{3}\operatorname{logit}(p_i^{(0)})\right).
\]
This is an independent per-question statistical correction; it does not use question text, reasoning traces, clusters, or consistency constraints. As Table~\ref{tab:platt-scaling-baseline} illustrates, Platt scaling is ineffective at improving either Brier score or consistency, making most metrics worse than the raw forecasts.

\begin{table}[H]
  \centering
  \caption{Fixed-parameter Platt scaling with coefficient $\sqrt{3}$.}
  \label{tab:platt-scaling-baseline}
  \begin{tabularx}{\textwidth}{@{}l*{4}{>{\centering\arraybackslash}X}@{}}
      \toprule
      Metric                        & Raw      & Platt    & $\Delta$           & Percent $\Delta$          \\
      \midrule
      Brier score                       & 0.151657 & 0.154199 & \mixed{$+0.002543$} & \mixed{$+1.68\%$} \\
      Negation squared error            & 0.046831 & 0.075584 & \bad{$+0.028753$}   & \bad{$+61.40\%$}   \\
      Multi-way partition squared error & 0.223655 & 0.212630 & \good{$-0.011025$}  & \good{$-4.93\%$}   \\
      Monotonicity $L_1$ violation      & 0.018565 & 0.019551 & \mixed{$+0.000986$} & \mixed{$+5.31\%$} \\
      \bottomrule
  \end{tabularx}
\end{table}

\subsection{Projection onto consistent solution set}
\label{app:projection-baselines}
We investigate whether our method is merely benefiting from the constraint of the probability axioms it is enforcing.
We can more simply apply these consistency constraints by minimizing the L2 deviation from the initial probabilities for each question while enforcing constraints. While projection enforces perfect consistency, the Brier improvement is minimal. Appendix~\ref{app:significance-testing} illustrates that this Brier improvement is statistically worse than that of our method.

\begin{table}[H]
  \centering
  \caption{Projection onto the consistent solution set with synthetic structural constraints enforced exactly.}
  \label{tab:projection-full-hard}
  \begin{tabularx}{\textwidth}{@{}l*{4}{>{\centering\arraybackslash}X}@{}}
      \toprule
      Metric                        & Raw      & Projection    & $\Delta$           & Percent $\Delta$          \\
      \midrule
      Brier score                       & 0.151657 & 0.147423 & \good{$-0.004233$} & \good{$-2.79\%$}   \\
      Negation squared error            & 0.046831 & 0.000000 & \good{$-0.046831$} & \good{$-100.00\%$} \\
      Multi-way partition squared error & 0.223655 & 0.000000 & \good{$-0.223655$} & \good{$-100.00\%$} \\
      Monotonicity $L_1$ violation      & 0.018565 & 0.000000 & \good{$-0.018565$} & \good{$-100.00\%$} \\
      \bottomrule
  \end{tabularx}
\end{table}

\subsection{Zero-shot LLM}
\label{app:zero-shot-llm}

Finally, we consider the zero-shot performance of GPT-5.1. Instead of using strict projection, we ask the same model used for the upstream forecaster, GPT-5.1, to update the probabilities given the consistency constraints. We use the following system prompt:

\begin{quote}
  \begin{Verbatim}[breaklines=true, breakanywhere=false,breaksymbolleft={}, breakautoindent=false, breakindent=0pt]
You are given a cluster of forecasting questions and an initial P(YES) for each question. Update all probabilities jointly using only the supplied question information and the explicit consistency constraints.

Reason about the questions as a coherent joint forecast. Reassess each initial probability in light of the event details, the other events in the cluster, and their logical relationships. Among probability assignments that satisfy every constraint, choose the values that best represent your substantive judgment of the events' likelihoods. The initial probabilities are estimates rather than fixed values; do not make arbitrary adjustments merely to force consistency.

The constraint block is authoritative:
- A complement constraint means the two listed probabilities must sum to 1.
- A MECE constraint means the listed mutually exclusive and collectively exhaustive probabilities must sum to 1.
- A monotonic chain P(Event A) <= P(Event B) means the probabilities must be non-decreasing from left to right.

Return one updated probability for every event, in the exact numbered event order. Satisfy all listed constraints simultaneously. Return only the required JSON object and no explanation.
  \end{Verbatim}
\end{quote}

The user prompt template is:

\begin{quote}
  \begin{Verbatim}[breaklines=true, breakanywhere=false,breaksymbolleft={}, breakautoindent=false, breakindent=0pt]
[Cluster of <N> forecasting events]

[Event 1]
Question: <question text>
Resolution date: <resolution date>
Background information: <background information>
Resolution criteria: <resolution criteria>
Initial predicted P(YES): <initial probability, rounded to four decimal places>

...

[Event N]
Question: <question text>
Resolution date: <resolution date>
Background information: <background information>
Resolution criteria: <resolution criteria>
Initial predicted P(YES): <initial probability, rounded to four decimal places>

[Consistency constraints]

[Complement constraints: each pair must sum to 1]
- P(Event <i>) + P(Event <j>) = 1

[MECE constraints: each group must sum to 1]
- P(Event <i>) + ... + P(Event <j>) = 1

[Monotonic chains: probabilities must be non-decreasing]
- P(Event <i>) <= ... <= P(Event <j>)

Return exactly <N> probabilities in Event 1 through Event <N> order.
  \end{Verbatim}
\end{quote}

We enforce the output to be a JSON object of the form \verb|{"probabilities": [...]}|.
We check that the array has length $N$ and had values in $[0,1]$ through downstream validation.

\begin{table}[H]
  \centering
  \caption{Zero-shot GPT-5.1 joint-reconciliation baseline on synthetic-question clusters.}
  \label{tab:zero-shot-llm-baseline}
  \begin{tabularx}{\textwidth}{@{}l*{4}{>{\centering\arraybackslash}X}@{}}
      \toprule
      Metric                            & Raw      & Zero-Shot GPT-5.1 & $\Delta$            & Percent $\Delta$          \\
      \midrule
      Brier score                       & 0.151657 & 0.157692          & \bad{$+0.006035$}  & \bad{$+3.98\%$}           \\
      Negation squared error            & 0.046831 & 0.000000          & \good{$-0.046831$} & \good{$-100.00\%$}        \\
      Multi-way partition squared error & 0.223655 & 0.000000          & \good{$-0.223655$} & \good{$-100.00\%$}        \\
      Monotonicity $L_1$ violation      & 0.018565 & 0.000000          & \good{$-0.018565$} & \good{$-100.00\%$}        \\
      \bottomrule
  \end{tabularx}
\end{table}

The zero-shot LLM satisfies all three constraint families up to numerical
precision, but worsens Brier score by $3.98\%$ relative to the raw forecasts.

\section{Additional ablation details}
\label{app:detailed-ablations}
In this section we provide detailed results for each ablation mentioned in Appendix~\ref{app:ablation-summary}, including per-seed metrics when applicable. The methodology for each ablation is described in Appendix~\ref{app:ablation-summary}, and significance testing is shown in Appendix~\ref{app:significance-testing}.

\paragraph{Ablation: no consistency loss}\leavevmode

\begin{table}[H]
  \centering
  \caption{Ablation removing the consistency term from the loss on synthetic-question clusters.}
  \label{tab:no-consistency-loss-synthetic-ablation}
  \begin{tabularx}{\textwidth}{@{}l*{4}{>{\centering\arraybackslash}X}@{}}
      \toprule
      Metric                        & Raw      & Ablation    & $\Delta$           & Percent $\Delta$          \\
      \midrule
      Brier score                       & 0.151657 & 0.136959 & \good{$-0.014697$} & \good{$-9.69\%$}  \\
      Negation squared error            & 0.046831 & 0.022303 & \good{$-0.024527$} & \good{$-52.37\%$} \\
      Multi-way partition squared error & 0.223655 & 0.074718 & \good{$-0.148938$} & \good{$-66.59\%$} \\
      Monotonicity $L_1$ violation      & 0.018565 & 0.011626 & \good{$-0.006939$} & \good{$-37.38\%$} \\
      \bottomrule
  \end{tabularx}
\end{table}

\begin{table}[H]
  \centering
  \caption{Per-seed breakdown of Table~\ref{tab:no-consistency-loss-synthetic-ablation}. Each column reports the metric value with its relative change from Raw in parentheses (negative is better).}
  \label{tab:no-consistency-loss-synthetic-ablation-per-seed}
  \begin{tabularx}{\textwidth}{@{}
    >{\hsize=1.12\hsize\raggedright\arraybackslash}X
    r
    *{3}{>{\hsize=.96\hsize\centering\arraybackslash}X}@{}}
      \toprule
      Metric                            & Raw      & Seed 31                         & Seed 32                          & Seed 33                         \\
      \midrule
      Brier score                       & 0.1517 & $0.1366$ \good{$(-9.93\%)$}  & $0.1368$ \good{$(-9.79\%)$}  & $0.1375$ \good{$(-9.36\%)$}  \\
      Negation sq.\ error               & 0.0468 & $0.0291$ \good{$(-37.81\%)$} & $0.0182$ \good{$(-61.16\%)$} & $0.0196$ \good{$(-58.15\%)$} \\
      Partition sq.\ error              & 0.2237 & $0.0506$ \good{$(-77.39\%)$} & $0.0684$ \good{$(-69.42\%)$} & $0.1052$ \good{$(-52.97\%)$} \\
      Monotonicity $L_1$ viol.          & 0.0186 & $0.0109$ \good{$(-41.06\%)$} & $0.0111$ \good{$(-40.12\%)$} & $0.0128$ \good{$(-30.96\%)$} \\
      \bottomrule
  \end{tabularx}
\end{table}

\paragraph{Ablation: no reasoning traces}\leavevmode

\begin{table}[H]
  \centering
  \caption{Ablation removing reasoning traces from prompts on synthetic-question clusters.}
  \label{tab:no-reasoning-traces-synthetic-ablation}
  \begin{tabularx}{\textwidth}{@{}l*{4}{>{\centering\arraybackslash}X}@{}}
      \toprule
      Metric                        & Raw      & Ablation    & $\Delta$           & Percent $\Delta$          \\
      \midrule
      Brier score                       & 0.151657 & 0.147610 & \good{$-0.004047$} & \good{$-2.67\%$}  \\
      Negation squared error            & 0.046831 & 0.007030 & \good{$-0.039801$} & \good{$-84.99\%$} \\
      Multi-way partition squared error & 0.223655 & 0.029545 & \good{$-0.194111$} & \good{$-86.79\%$} \\
      Monotonicity $L_1$ violation      & 0.018565 & 0.003830 & \good{$-0.014735$} & \good{$-79.37\%$} \\
      \bottomrule
  \end{tabularx}
\end{table}

\begin{table}[H]
  \centering
  \caption{Per-seed breakdown of Table~\ref{tab:no-reasoning-traces-synthetic-ablation}. Each column reports the metric value with its relative change from Raw in parentheses (negative is better).}
  \label{tab:no-reasoning-traces-synthetic-ablation-per-seed}
  \begin{tabularx}{\textwidth}{@{}
    >{\hsize=1.12\hsize\raggedright\arraybackslash}X
    r
    *{3}{>{\hsize=.96\hsize\centering\arraybackslash}X}@{}}
      \toprule
      Metric                            & Raw      & Seed 31                        & Seed 32                        & Seed 33                        \\
      \midrule
      Brier score                       & 0.1517 & $0.1473$ \good{$(-2.85\%)$}  & $0.1476$ \good{$(-2.70\%)$}  & $0.1479$ \good{$(-2.46\%)$}  \\
      Negation sq.\ error               & 0.0468 & $0.0069$ \good{$(-85.16\%)$} & $0.0073$ \good{$(-84.50\%)$} & $0.0069$ \good{$(-85.30\%)$} \\
      Partition sq.\ error              & 0.2237 & $0.0306$ \good{$(-86.30\%)$} & $0.0220$ \good{$(-90.18\%)$} & $0.0360$ \good{$(-83.90\%)$} \\
      Monotonicity $L_1$ viol.          & 0.0186 & $0.0036$ \good{$(-80.35\%)$} & $0.0038$ \good{$(-79.78\%)$} & $0.0041$ \good{$(-77.98\%)$} \\
      \bottomrule
  \end{tabularx}
\end{table}

\paragraph{Ablation: no cross-question context}\leavevmode

\begin{table}[H]
  \centering
  \caption{No cross-question context ablation that uses clusters containing ten independent runs of the same question, as opposed to clusters with related questions.}
  \label{tab:same-question-cluster-ablation}
  \begin{tabularx}{\textwidth}{@{}l*{4}{>{\centering\arraybackslash}X}@{}}
      \toprule
      Metric                        & Raw      & Ablation    & $\Delta$           & Percent $\Delta$          \\
      \midrule
      Brier score                       & 0.151657 & 0.138197 & \good{$-0.013460$} & \good{$-8.88\%$}  \\
      Negation squared error            & 0.046831 & 0.013080 & \good{$-0.033750$} & \good{$-72.07\%$} \\
      Multi-way partition squared error & 0.223655 & 0.095183 & \good{$-0.128473$} & \good{$-57.44\%$} \\
      Monotonicity $L_1$ violation      & 0.018565 & 0.009324 & \good{$-0.009241$} & \good{$-49.78\%$} \\
      \bottomrule
  \end{tabularx}
\end{table}

\begin{table}[H]
  \centering
  \caption{Per-seed breakdown of Table~\ref{tab:same-question-cluster-ablation}. Each column reports the metric value with its relative change from Raw in parentheses (negative is better).}
  \label{tab:same-question-cluster-ablation-per-seed}
  \begin{tabularx}{\textwidth}{@{}
    >{\hsize=1.12\hsize\raggedright\arraybackslash}X
    r
    *{3}{>{\hsize=.96\hsize\centering\arraybackslash}X}@{}}
      \toprule
      Metric                            & Raw      & Seed 31                        & Seed 32                        & Seed 33                        \\
      \midrule
      Brier score                       & 0.1517 & $0.1370$ \good{$(-9.64\%)$}  & $0.1391$ \good{$(-8.25\%)$}  & $0.1384$ \good{$(-8.74\%)$}  \\
      Negation sq.\ error               & 0.0468 & $0.0147$ \good{$(-68.71\%)$} & $0.0126$ \good{$(-73.16\%)$} & $0.0120$ \good{$(-74.33\%)$} \\
      Partition sq.\ error              & 0.2237 & $0.0979$ \good{$(-56.21\%)$} & $0.0923$ \good{$(-58.71\%)$} & $0.0953$ \good{$(-57.41\%)$} \\
      Monotonicity $L_1$ viol.          & 0.0186 & $0.0096$ \good{$(-48.28\%)$} & $0.0094$ \good{$(-49.28\%)$} & $0.0090$ \good{$(-51.77\%)$} \\
      \bottomrule
  \end{tabularx}
\end{table}

\paragraph{Ablation: frontier LLM}\leavevmode

\begin{table}[H]
  \centering
  \caption{Ablation replacing the finetuned model with GPT-5.1 off-the-shelf.}
  \label{tab:no-finetuning-synthetic-ablation}
  \begin{tabularx}{\textwidth}{@{}l*{4}{>{\centering\arraybackslash}X}@{}}
      \toprule
      Metric                            & Raw      & GPT-5.1  & $\Delta$           & Percent $\Delta$          \\
      \midrule
      Brier score                       & 0.151657 & 0.142821 & \good{$-0.008836$} & \good{$-5.83\%$}  \\
      Negation squared error            & 0.046831 & 0.036621 & \good{$-0.010210$} & \good{$-21.80\%$} \\
      Multi-way partition squared error & 0.223655 & 0.180908 & \good{$-0.042748$} & \good{$-19.11\%$} \\
      Monotonicity $L_1$ violation      & 0.018565 & 0.018164 & \good{$-0.000401$} & \good{$-2.16\%$}  \\
      \bottomrule
  \end{tabularx}
\end{table}

\end{document}